\documentclass[10pt,twocolumn,letterpaper]{article}

\pdfoutput = 1
\usepackage[pagenumbers]{cvpr} 

\usepackage{cite}
\usepackage{algorithm}
\usepackage{algorithmic}
\usepackage{graphicx}
\usepackage{cite}
\usepackage{caption}
\usepackage{epigraph}
\usepackage[switch]{lineno}
\usepackage{mathtools}
\usepackage{multirow}
\usepackage{multicol}
\usepackage{amsmath,amssymb,amsfonts}
\usepackage{graphicx}
\usepackage{xcolor}
\usepackage{url}
\usepackage{amsmath}

\usepackage{diagbox}

\usepackage{rotating}

\definecolor{cvprblue}{rgb}{0.21,0.49,0.74}
\usepackage[pagebackref,breaklinks,colorlinks,allcolors=cvprblue]{hyperref}

\def\paperID{*****} 
\def\confName{CVPR}
\def\confYear{2025}

\title{Continual Unlearning for Foundational Text-to-Image Models without Generalization Erosion}

\author{Kartik Thakral$^1$, Tamar Glaser$^2$, Tal Hassner$^3$, Mayank Vatsa$^1$, Richa Singh$^1$ \\
$^1$IIT Jodhpur, $^2$Harman International, $^3$Weir AI\\
{\tt\small \{thakral.1, mvatsa, richa\}@iitj.ac.in}, \tt\small \{tamarglasr, talhassner\}@gmail.com
}

\begin{document}

\twocolumn[{
\maketitle
\begin{center}
    \captionsetup{type=figure}
    \vspace{-14pt}
    \includegraphics[width=0.75\textwidth]{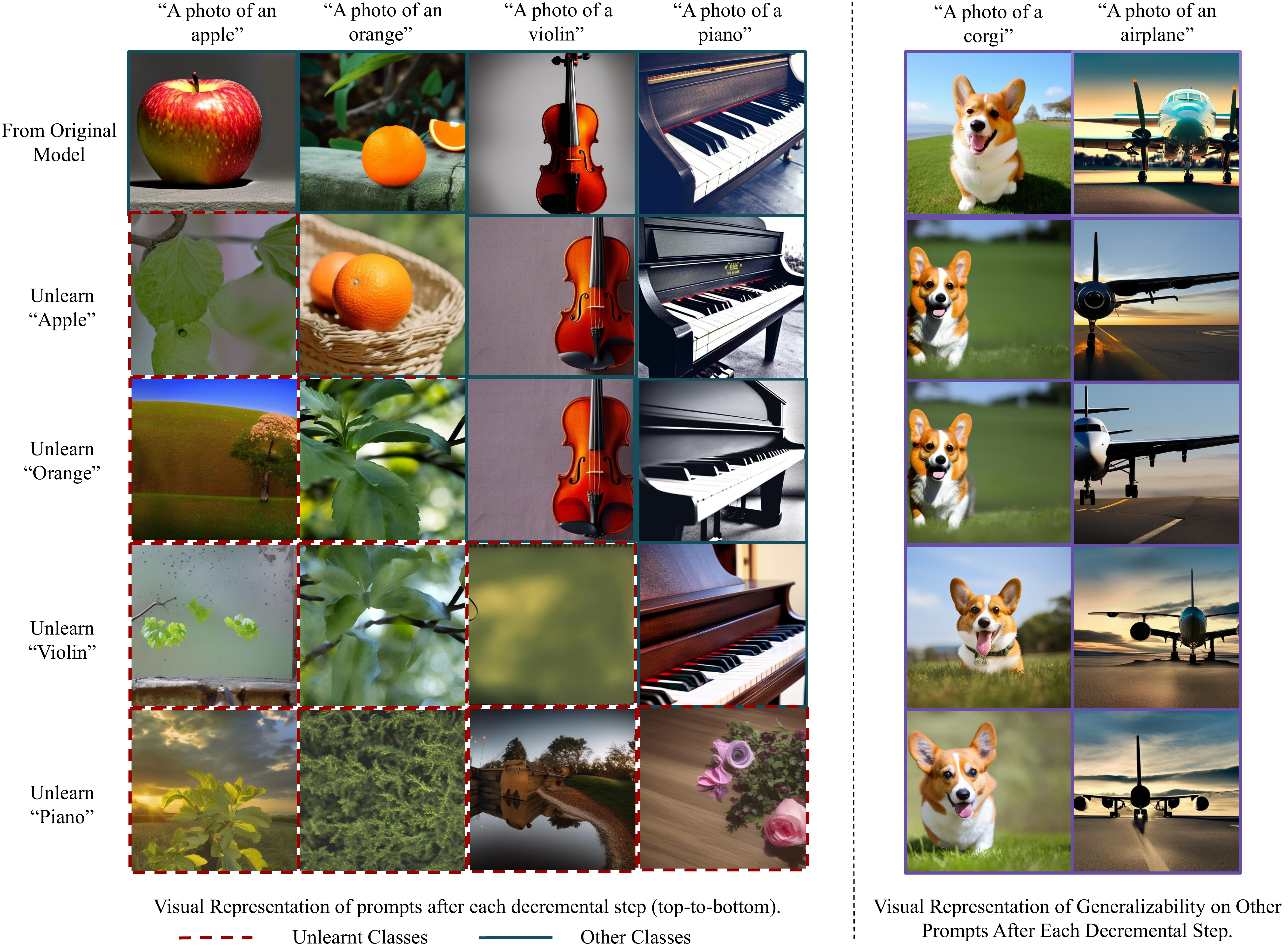}
    \vspace{-3pt}
    \captionof{figure}{Continual Unlearning: Illustrating the methodology for selective forgetting in text-to-image foundational generative models like Midjourney, stable diffusion, and Imagen. This illustration outlines the multi-step process for the removal of specific concepts from the model's knowledge while simultaneously ensuring that its ability to generalize over other concepts remains intact.}
\end{center}
}]

\begin{abstract}

\vspace{-10pt}

How can we effectively unlearn selected concepts from pre-trained generative foundation models without resorting to extensive retraining? This research introduces `continual unlearning', a novel paradigm that enables the targeted removal of multiple specific concepts from foundational generative models, incrementally. We propose Decremental Unlearning without Generalization Erosion (DUGE) algorithm which selectively unlearns the generation of undesired concepts while preserving the generation of related, non-targeted concepts and alleviating generalization erosion. For this, DUGE targets three losses: a cross-attention loss that steers the focus towards images devoid of the target concept; a prior-preservation loss that safeguards knowledge related to non-target concepts; and a regularization loss that prevents the model from suffering from generalization erosion. Experimental results demonstrate the ability of the proposed approach to exclude certain concepts without compromising the overall integrity and performance of the model. This offers a pragmatic solution for refining generative models, adeptly handling the intricacies of model training and concept management lowering the risks of copyright infringement, personal or licensed material misuse, and replication of distinctive artistic styles. Importantly, it maintains the non-targeted concepts, thereby safeguarding the model's core capabilities and effectiveness.
\end{abstract}    
\section{Introduction}
\label{sec:intro}
\par Recent advances in large-scale text-to-image models have been notable for synthesizing highly realistic images \citep{ramesh2022hierarchical, nichol2021glide, saharia2022photorealistic, rombach2022high}. However, the training process for these models is resource-intensive, requiring extensive datasets, GPUs, and long periods of computation time. For instance, the training of Stable Diffusion on a subset of 2 billion images from the LAION-5B dataset \citep{schuhmann2022laion} required about 250 hours using 256 GPUs. Training Imagen on the entire LAION-5B dataset took nearly 750 hours on 512 TPUs. The significant costs and environmental impact of training these foundational generative models are a growing concern. Furthermore, these training datasets, often sourced from the web \citep{schuhmann2022laion}, may include copyrighted material, artistic works, and personal images \citep{somepalli2023diffusion, carlini2023extracting, shan2023glaze, Imagen, mishkin2022dall, rando2022red} that the models learn and generate indiscriminately. This indiscriminate learning also extends to false information \citep{boswald2022pixel, fairobserver, xu2023exposing}, which can be spread through the generated images.

\par With the emergence of AI regulations like CCPA \citep{ccpa} and GDPR \citep{voigt2017eu}, creators should be able to get their images removed from these models. However, retraining models for each removal is not a viable option due to computational constraints. This research, therefore, focuses on effective continual unlearning to facilitate the forgetting of specific concepts in pre-trained text-to-image models \citep{SDM, rombach2022high}. We are guided by three crucial questions: \textit{(i) How can we ensure text-to-image models forget certain concepts without impairing their other functions? (ii) How can these models continually unlearn multiple concepts while maintaining their overall performance? (iii) How to assess the effectiveness of incremental unlearning in these models?}

\par To explore these research questions, we begin by defining the concept of continual unlearning in the context of text-to-image foundational generative models as: \textit{``Continual unlearning is the process of systematically and iteratively removing the knowledge of specific concepts from a generative model's knowledge, effectively preventing it from synthesizing the previously unlearned concepts’’}. Since the objective is to forget multiple concepts from the model continually, the process is also interchangeably referred to as \textit{incremental unlearning}. 

\par While current research has examined algorithms for correcting facts, updating concepts, and concept ablation \citep{arad2023refact, avrahami2022blended, zhang2023sine, gandikota2023erasing, kumari2023ablating}, none have discussed forgetting multiple concepts from a generative model. This paper addresses the gap by examining the complex issue of continual unlearning in text-to-image generative models. We also observe an interesting effect: as these models undergo incremental unlearning, they tend to exhibit a decline in their ability to generate images for various prompts. We refer to this phenomenon as ``generalization erosion’’, akin to the catastrophic forgetting observed in incremental learning of discriminative models. In addition, we propose efficient methods to assess the degree of continual unlearning in these models, both qualitatively and quantitatively.

\par To address the challenging task of continual unlearning in generative models, we introduce \textit{DUGE: Decremental Unlearning without Generalization Erosion}, a memory-centric algorithm designed to enable unlearning at each decremental step. This is accomplished by strategic intervention in the cross-attention layers of the model while concurrently applying regularization techniques to retain the model’s capability to generalize across unrelated concepts. The key contributions of this paper are as follows:
\begin{enumerate}
    \item Canonicalization of continual unlearning concept in text-to-image foundational generative models.

    \item Introduction of ``Generalization Erosion,’’ a newly observed phenomenon in text-to-image generative models as a result of continual unlearning.

    \item Formulation of \textit{DUGE}, a novel memory-based algorithm tailored explicitly for decremental learning in text-to-image generative models.

\end{enumerate}

\par Experiments on our Dec-ImageNet-20 and MS-COCO datasets demonstrate that the algorithm successfully unlearns multiple concepts while retaining its ability to generalize concepts unrelated to the unlearning process. Furthermore, a user study shows an average performance of over 95\% for continual unlearning. This highlights the algorithm's effectiveness in continual unlearning and its resilience against the phenomenon of generalization erosion.

\section{Related Work}
Recent advancements in machine learning have focused on the intertwined methods of concept editing and forgetting. Incremental learning is pivotal here, as it investigates how models can effectively assimilate new knowledge while preserving previously acquired concepts. Decremental learning, on the other hand, focuses on the intentional forgetting of target concepts without losing the generalizability over prior knowledge. The integration of both, incremental and decremental is instrumental in the progression of machine~learning.

\noindent\textbf{Continual Learning:}
This focuses on creating models capable of continuously acquiring new concepts while retaining previously learned ones to address the challenge of catastrophic forgetting \citep{mccloskey1989catastrophic, nguyen2019toward}. In this domain, class incremental learning involves introducing new classes sequentially to models, aiming to learn these new classes without compromising performance on existing ones. Class incremental learning approaches are broadly categorized into three groups: (i) regularization-based methods, which limit weight modifications to preserve knowledge \citep{petit2023fetril, zhou2021co}, (ii) dynamic architecture techniques, which adapt the model structure to accommodate new information \citep{yoon2019scalable, sarwar2019incremental}, and (iii) replay or generative replay mechanisms, designed to periodically revisit old tasks for reinforcement \citep{ji2022complementary, cha2021co2l}.

Recently, class-incremental learning has achieved notable advancements in computer vision, including self-supervised approaches \citep{thakral2023self, kalla2022s3c}, which leverage unlabeled data, generative samplers and attention distillation methods for image classification \citep{liu2020generative, sajedi2023datadam}, and innovative prompt-based algorithms \citep{wang2022dualprompt, smith2023coda, wang2022learning}. The ongoing progress in this field presents a promising direction for the development of adaptive systems to assimilate new visual knowledge.

\noindent\textbf{Generative Machine Unlearning:} Concept editing and concept forgetting are two primary research threads in this domain. Foundational generative models, trained on extensive datasets, often lack mechanisms for updating factual information over time. To address this, researchers have developed algorithms to modify specific details in the model’s knowledge, a process referred to as \textit{Concept Editing}. 

\par Avrahami et al. \citep{avrahami2022blended} proposed text-driven editing of natural images using a combination of a pre-trained diffusion and CLIP model. This method allows for localized changes based on text prompts while maintaining the integrity of the rest of the image. Zhang et al. \citep{zhang2023sine} introduced a single-image editing technique using text-to-image diffusion models, which fine-tunes a pre-trained diffusion model with a single image and a text descriptor. Wallace et al. \citep{wallace2023edict} proposed accurately reversing the diffusion process in diffusion models using a pair of interrelated noise vectors for precise recovery of the original noise vector from real or model-generated images. Mokady et al. \citep{mokady2023null} developed a dual-stage process for editing real images with text-guided diffusion models, combining DDIM sampling with null-text optimization for intuitive text-based image editing without requiring model adjustments or mask inputs. Arad et al. \citep{arad2023refact} proposed updating factual content in text-to-image models by altering a specific layer in the text encoder, ensuring that image quality and unrelated facts remain unaffected. 

These concepts can also be applied to Generative Adversarial Networks (GANs). Bau et al. \citep{bau2020rewriting} proposed adding or editing a specific concept within the associative memory \citep{kohonen1973representation} of a GAN as a constrained optimization problem, extending associative memory to the nonlinear convolutional layers. Nobari et al. \citep{nobari2021creativegan} and Wang et al. \citep{wang2022rewriting} developed methods to enhance the knowledge base of GANs. They aimed to prevent mode collapse, thereby fostering greater creativity in the output of these networks.

Researchers have approached the task of machine unlearning in generative models with the aim of forgetting specific learned information or erasing the influence of a particular subset of training data from a trained model. This process is known as \textit{Concept Forgetting}. Although there has been significant progress in unlearning in discriminative models \citep{sekhari2021remember,gupta2021adaptive,tarun2023fast,liu2023muter}, unlearning in generative models is relatively new and less explored. Gandikota et al. \citep{gandikota2023erasing} developed a method to erase specific visual concepts from text-to-image diffusion models by using negative guidance for fine-tuning model weights, which has proven effective in removing explicit content, artistic styles, and object classes. Kumari et al. \citep{kumari2023ablating} proposed concept ablation in text-to-image diffusion models via concept replacement, which uses KL Divergence to match the image distribution of a target concept to an anchor concept. In GANs, Tiwari et al. \citep{tiwary2023adapt} proposed a two-step unlearning process, in which the network is first adapted to the concepts to be forgotten, and then the original network is trained only on the desirable concepts while keeping the weights distinct from those of the adapted model.

Previous research has predominantly concentrated on incremental learning of concepts or revising factual knowledge and ablating concepts from generative and discriminative models. However, the nuanced challenges associated with incremental unlearning in foundational generative models have not been thoroughly examined in a formal context. To the best of our knowledge, this study is the first to articulate the issue of continual unlearning in generative models and to develop methods specifically for progressively eliminating multiple concepts from text-to-image generation systems. The proposed algorithm facilitates effective unlearning of specific concepts without the need for complete model retraining.

\begin{figure*}[t]
\centering
    \includegraphics[scale=0.07]{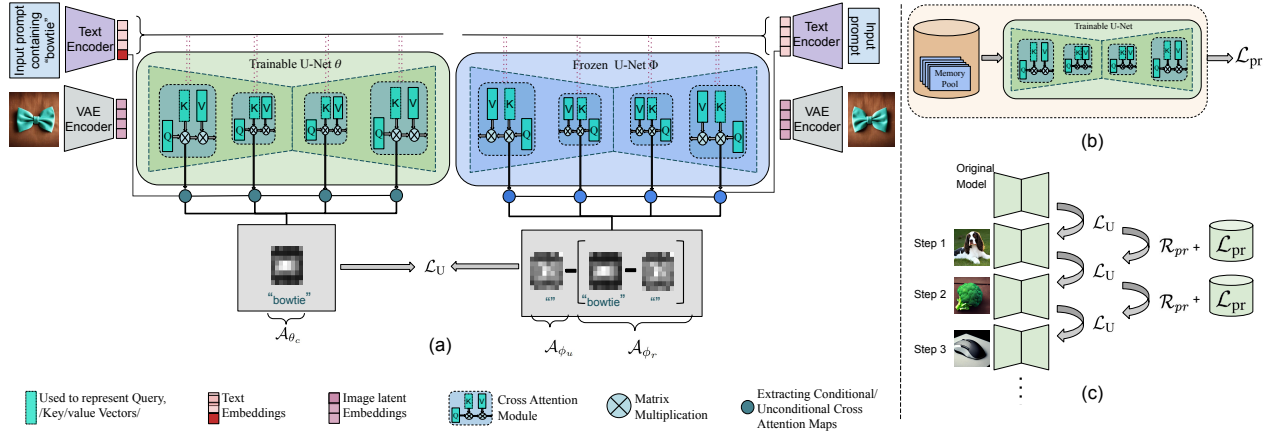}
    \caption{Illustration of the complete DUGE algorithm for the continual unlearning process. (a) Depicts the unlearning process employed at each decremental step; (b) depicts the prior-preservation loss utilizing a small memory for regularization; and (c) demonstrates the complete decremental framework of the DUGE formulation with multiple decremental steps.}

\label{fig:methodology}
\end{figure*}

\section{Background}
In this paper, we focus on foundational text-to-image generative models. We begin with a brief overview of text-to-image diffusion models in Section \ref{sec:background}. Subsequently, a comprehensive description of the continual unlearning is given in Section \ref{sec:prob_description}. The proposed \textit{DUGE} algorithm for decremental learning is elaborated in two parts (in Section \ref{sec:unlearning}): (i) the process of unlearning during each decremental step, and (ii) the application of regularization techniques specific to decremental steps.

\subsection{Preliminaries and Definitions}
\label{sec:background}

We first review the standard terminology for diffusion models, which is later used in this paper. These models learn the distribution space through a step-by-step denoising process \citep{ho2020denoising, SDM}. They start with Gaussian noise and gradually denoise over $T$ steps to form a final image. The model predicts noise $\mathcal{\epsilon}_t$ at each step $t$, generating the intermediate denoised image $x_t$. Here, $x_T$ is the initial noise and $x_0$ is the final image. This process is modeled as a Markov Transition~Probability. 

\begin{equation}
    p_\theta(x_T:0) = p(x_T) \prod_{t=T}^1 p_\theta(x_{t-1}|x_t)
\end{equation}

Latent Diffusion Models (LDMs) \citep{rombach2022high} are commonly employed in text-to-image foundational generative models, enhancing efficiency by functioning within a lower-dimensional latent space $z$. This space is part of a pre-trained variational autoencoder, equipped with an encoder $\mathcal{E}$ and a decoder $\mathcal{D}$. In the training phase, an image $x$ undergoes noise addition in its encoded latent, resulting in $z = \mathcal{E}(x)$, which subsequently leads to $z_t$, where the noise level intensifies with $t$. The LDM process can be viewed as a series of denoising models with identical parameters $\theta$. These models are trained to predict the noise $\epsilon_\theta (z_t, c, t)$ that is added to $z_t$, conditioned on both the timestep $t$ and a text condition $c$. The objective function for optimization is as~follows.
\begin{equation}
    \mathcal{L} = \mathbb{E}_{z_t \in \mathcal{E}(x), t, c, \epsilon \sim \mathcal{N}(0,1)}[\|\epsilon - \epsilon_\theta(z_t, c, t)\|_2^2]
\end{equation}

\subsection{Continual Unlearning}
\label{sec:prob_description}

Continual Unlearning is a challenging problem because it requires modifying the pre-trained model parameters without compromising its original performance on unrelated concepts. To address this, we utilize a dataset organized for incremental unlearning where each $D^\delta$ contains $N^\delta$ image-text pairs.
\begin{equation}
D = \{D^\delta\}_{\delta=1}^\Delta
\end{equation}
 \noindent \textit{Decremental set} $\mathcal{S}=\{s_1, s_2,...,s_\Delta\}$  specifies the concepts to be removed from the model's knowledge for each decremental task $\delta \in [1, \Delta]$. We train the model $\epsilon_\theta$ decrementally for each task $\delta$, resulting in a final model $\epsilon_\theta^\Delta$ that is unable to generate images with concepts from the decremental set~$\mathcal{S}$.

\section{Proposed Approach}
\label{sec:unlearning}
We view continual unlearning as the challenge of keeping the model close to its previous version while performing unlearning at each decremental step. To unlearn a target concept, the proposed \textit{DUGE} algorithm aims to distort the alignment between the cross-attention of image and text embeddings. Sequentially forgetting concepts can lead to a degradation in the quality of generation from other prompts. To address the generalization erosion of different prompts, DUGE utilizes a small memory bank to regularize the model and keep it bounded around the same initial decremental steps. This strategy ensures successful unlearning of the target concept while mitigating generalization erosion for different prompts. 

\subsection{Unlearning} 
Text-to-image diffusion models are primarily guided by the weights of cross-attention layers to align the image generation process with the provided text input. To unlearn a specific concept from a pre-trained diffusion model, we aim to disrupt this alignment between a target class and the generated output. This can be done by regulating the attention probabilities associated with a target class, as it is the key component within a cross-attention layer. 

Consider a pre-trained diffusion model, parameterized by $\theta$, which is to be fine-tuned, and a corresponding frozen diffusion model, parameterized by $\phi$. Let $\mathcal{A}_\theta$ and $\mathcal{A}_\phi$ represent their respective attention maps. We denote the attention-probability maps for a total of $l$ cross-attention layers as $\mathcal{A}_{\theta} = \{a_i\}_{i=1}^{l}$. For our primary objective, we define two types of cross-attention maps: (i) Context-conditional attention maps, represented as $\mathcal{A}_{\phi_c}$, are the attention probability maps extracted for a specific target concept to be unlearned in the input text. (ii) Unconditional cross-attention maps ($\mathcal{A}_{\phi_u}$), are the cross-attention maps generated when an empty prompt ``'' is~provided.

Since the task is to deviate $\mathcal{A}_{\theta_c}$ from the parts of the image most relevant to the target concept, we first calculate the residual cross-attention maps $\mathcal{A}_{\phi_r}$ by subtracting $\mathcal{A}_{\phi_u}$ from $\mathcal{A}_{\phi_c}$ highlighting the most relevant parts of the image with respect to target concept. This is because for a target concept to be unlearnt, $\mathcal{A}_{\phi_u}$ does not highlight any specific portion of the image. Once $\mathcal{A}_{\phi_r}$ is obtained, we can now obtain the desired part of the cross-attention maps least relevant with respect to the target concept by subtracting $\mathcal{A}_{\phi_r}$ from $\mathcal{A}_{\phi_u}$. We define it as $\mathcal{A}_{\phi_n}$ representing negative cross-attention maps, highlighting the portions of the maps when a concept is not provided in the input text.
\begin{equation}
 \mathcal{A}_{\phi_n} = \mathcal{A}_{\phi_u} - (\mathcal{A}_{\phi_c} - \mathcal{A}_{\phi_u})    
\end{equation}

\noindent The final unlearning loss function $\mathcal{L}_{\text{U}}$ is then  formulated by taking the mean squared error between  $\mathcal{A}_{\theta_c}$ and $\mathcal{A}_{\phi_n}$. 
\begin{equation}
    \mathcal{L}_{\text{U}} = \frac{1}{M} \sum_{i=1}^{M} (\mathcal{A}_{\theta_c}^i - \mathcal{A}_{\phi_n}^i)^2
\end{equation}
Here, $M$ represents the total number of elements in the cross-attention maps. By modifying them during the retraining phase, the model is constrained to generate images that exclude the target concept. For the decremental step $\delta$, this effectively results in the model unlearning the target concept $s_\delta$.

\subsection{Memory Regularization}
\label{sec:dec_learning}
We observe that when text-to-image generative models are finetuned decrementally, they tend to lose generalization over other prompts. In other words, the semantic space of the decrementally finetuned models significantly degrades leading to \textit{Generalization Erosion}, i.e., either the generated output is semantically not coherent with the input prompt or the visual quality of the generated images is significantly degraded. An ideal decremental learning algorithm would not only unlearn the target concept but also maintain its performance on other prompts, displaying its robustness against generalization erosion. To address this, DUGE utilizes a small memory $\mathcal{M} = (x_i, y_i)$ of size $m$ to regularize the weights $\theta^\delta$ to stay within the bounds of the original/starting model. This memory is curated from LAION-Aesthetics V2 dataset \citep{LAION-aesthetics} while ensuring ${y}_{i=1}^m \notin \mathcal{S}$. We utilize only the captions from this dataset and generate data $z^*_{t_i} = \epsilon(\mathcal{E}(x_i))$ by using the ancestral sampler on the frozen pre-trained diffusion model with random initial noise $z_{t_i} \sim \mathcal{N}(0, I)$ and conditioning vector $c_{pr} = \text{(``a [class\_noun]”)}$. A prior-preservation loss is minimized on the samples of the memory to prevent generalization erosion and overfitting \citep{ruiz2023dreambooth} described below:
\begin{equation}
    \mathcal{L}_{\text{pr}} = \frac{1}{m} \sum_{i=1}^{m} \mathbb{E}_{x_t \in \mathcal{E}(x), x \in \mathcal{M}, t, c_{pr}, \epsilon \sim \mathcal{N}(0,1)}[\|\epsilon - \epsilon_{\theta^\delta}(x, c_{pr}, t)\|_2^2]
\end{equation}

\noindent In addition to the optimization of $\mathcal{L}_{\text{pr}}$, we also regularize the weights of the finetuned model at $\delta$ decremental step with a Kullback–Leibler (KL) divergence penalty with the model of $\delta - 1$ step \citep{xuhong2018explicit, nguyen2019toward} described below:

\begin{equation}
    \mathcal{R}_{pr} = \min_{\theta^\delta} \mathcal{D_{KL}} (\theta^\delta || \theta^{\delta-1})
\end{equation}

\noindent Here, model weights at steps $\delta - 1$ and $\delta$ are treated as two distinct distributions and are optimized to be bounded to prevent generalization erosion on non-targeted prompts by keeping models close to their predecessors. \noindent For training the final model $\theta^\delta$, the three losses are jointly minimized with $\lambda_i$ as the relative weight of each loss.

\begin{equation}
    \mathcal{L}_{\text{Total}} = \min_{\theta^\delta}  \lambda_1 \mathcal{L}_{\text{U}} + \lambda_2 \mathcal{L}_{pr} + \lambda_3 \mathcal{R}_{pr}
\end{equation}

\section{Experimental Details}
This section provides various intricate details of the experiments performed around decremental learning.

\begin{table}[b]
\centering
\caption{Total number of clean prompts used for generalization erosion evaluations for each set after removing all the class names in the validation set of the MS-COCO dataset.}
\begin{tabular}{lllll}
\hline
               & Set 1 & Set 2 & Set 3 & Set 4 \\ \hline
No. of Prompts & 4844  & 4924  & 4992  & 4838  \\ \hline
\end{tabular}

\label{tab:prompt_count}
\end{table}

\subsection{Dataset, Evaluation Protocol, and Metrics}
For continual unlearning, we define the protocols for 3 and 5 decremental steps. Toward this end, we identify inter-related object classes and create a subset of 20 classes from ImageNet-1K dataset \citep{deng2009imagenet} comprising a total of 4 sets as described in Table \ref{tab:dec_imagenet_sets}. Classes are taken from the intersection of ImageNet-1K and MS-COCO to evaluate the corresponding captions. We term this as the Dec-ImageNet-20 dataset. For unlearning, we utilize the knowledge of the pre-trained source model only. For a target concept to be unlearnt, we take four prompts mentioning the target concept and its corresponding four generated output images. To evaluate for generalization, we use the validation set with 5k images of the MS-COCO dataset \citep{lin2014microsoft}. We exclude the prompts mentioning the direct class reference to evaluate each decremental set for generalization erosion, resulting in around 4.8K prompts for each set. The details of the total number of prompts for each set are provided in Table \ref{tab:prompt_count}.

\begin{table*}[t]
\caption{Details of the replacement classes used for Naive-Finetuning. The replacements are randomly selected for each target class from the ImageNet dataset.} 
\centering
\resizebox{\textwidth}{!}{%
\begin{tabular}{cccccccc}
\hline
\multicolumn{2}{c}{Set 1}                                                                                            & \multicolumn{2}{c}{Set 2} & \multicolumn{2}{c}{Set 3}                                                 & \multicolumn{2}{c}{Set 4}                                                                                              \\ \hline
Original                                                 & Replacement                                               & Original   & Replacement  & Original                                                    & Replacement & Original                                                 & Replacement                                                 \\ \hline
Apple                                                    & Banana                                                    & Goldfish   & Broom        & \begin{tabular}[c]{@{}c@{}}Gas Pump\end{tabular}         & Broom       & \begin{tabular}[c]{@{}c@{}}Golf Ball\end{tabular}     & Parachute                                                   \\
Broccoli                                                 & Sunglass                                                  & Bowtie     & Mango        & Sunglass                                                    & Mango       & Pizza                                                    & \begin{tabular}[c]{@{}c@{}}Computer Keyboard\end{tabular} \\
Backpack                                                 & Mango                                                     & Harp       & Banana       & \begin{tabular}[c]{@{}c@{}}Computer Mouse\end{tabular}   & Harp        & Tench                                                    & Horse                                                       \\
\begin{tabular}[c]{@{}c@{}}Traffic Light\end{tabular} & \begin{tabular}[c]{@{}c@{}}Computer Mouse\end{tabular} & Parachute  & Sunglass     & Broom                                                       & Orange      & \begin{tabular}[c]{@{}c@{}}Garbage Truck\end{tabular} & Banana                                                      \\
Umbrella                                                 & Horse                                                     & Orange     & Umbrella     & \begin{tabular}[c]{@{}c@{}}English Springer\end{tabular} & Umbrella   & \begin{tabular}[c]{@{}c@{}}Teddy Bear\end{tabular}    & Sunglass                                                    \\ \hline
\end{tabular}
}
\label{tab:replacement}
\end{table*}

\noindent\textbf{Evaluation Protocol and Metrics:} In this work, it is assumed that a concept is characterized by a “label” i.e  as class noun. For each decremental step $\delta$, we start with the unlearning of the class $s_i$ in a set and report the class-wise accuracy for each class of the set with 500 images generated with prompt ``A photo of a \textit{$<$class-name$>$}''. A pre-trained ResNet-50 \citep{he2016deep} is utilized to report the classification performance. To evaluate for generalization erosion at each step $\delta$, we first generate the outputs for evaluation prompt dataset using the source model and then report FID (Fréchet Inception Distance) \citep{heusel2017gans} and KID (Kernel Inception Distance) \citep{dai2021knowledge} between the generated outputs of $\delta$ unlearnt and source model. Stable Diffusion V2 is used as the source for the decremental setup and carries on for each step of the experiment. For each decremental step, lower class accuracy of the target class to be unlearnt represents better unlearning, and higher class-wise accuracy (closer to the original) of classes yet to be unlearnt represents lower generalization erosion. Lower FID and KID between the current and source models represent better generalizability over non-targeted concepts. In addition to analyzing whether the proposed algorithm can unlearn a concept, we also conduct a user study on set 1 and report the unlearning results.



\subsection{Implementation Details}
The official source code for Stable DiffusionV2\footnote{\url{https://huggingface.co/blog/stable_diffusion}} of the diffusers library\footnote{\url{https://huggingface.co/docs/diffusers/index}} is used for all experiments as the source model for the decremental setup. For reproducibility, we make the source and trained model publicly available\footnote{Project Website: \url{https://www.iab-rubric.org/unlearning-continual-unlearning}}.

\noindent\textbf{Naive-Finetuning:} As the baseline experiment, we first begin with fine-tuning the source model and proceed in a decremental fashion for the classes of each set defined in Table \ref{tab:dec_imagenet_sets}. For Naive-Finetuning, unlearning at each decremental step is performed by connecting the text prompt (mentioning the target class name) with the images of a random class (selected from a set of classes) generated by the source model. The details of the replacement classes chosen for the target class in each set are available in Table \ref{tab:replacement}.

\noindent\textbf{DUGE:} For training DUGE, we use 3 textual prompts per class that explicitly mention the class name to be unlearnt, along with the 5 images generated by the source model that best match each textual prompt. While training, $\lambda_3$ is set to 0.01, and we utilize an iteration-based lambda scheduler for $\lambda_1$ and $\lambda_2$. DUGE is trained for two settings, one with 50 iterations and the other with 35 iterations, along with a learning rate set to 2e-06 for all classes of each set, and the best of the two is considered. For data generation from the source model, we fixed the settings for the guidance scale to 7.0, the number of inference steps to 50, and the number of images per prompt to 2 for generalization erosion and 500 for classification accuracy evaluations.

\begin{table}[t]
\centering
\caption{List of classes in each set of the Dec-ImageNet-20 dataset. These classes are used to evaluate the proposed DUGE algorithm over 3 and 5 decremental steps.}
\resizebox{\linewidth}{!}{
\begin{tabular}{ccccc}
\hline
\begin{tabular}[c]{@{}c@{}}Decremental \\ Steps\end{tabular}  & Set 1         & Set 2     & Set 3            & Set 4         \\ \hline
Step 1            & Apple         & Goldfish  & Gas Pump         & Golf Ball     \\
Step 2            & Broccoli      & Blowtie   & Sunglass         & Pizza         \\
Step 3            & Backpack      & Harp      & Computer Mouse   & Trench        \\ \hline
Step 4            & Traffic Light & Parachute & Broom            & Garbage Truck \\
Step 5            & Umbrella      & Orange    & English Springer & Teddy Bear    \\ \hline
\end{tabular}}
\label{tab:dec_imagenet_sets}
\end{table}

\noindent\textbf{Generalization Erosion.} To experiment for generalization erosion, we chose the validation set of the MS-COCO dataset with 5k prompts. For each set, we exclude prompts mentioning the class names. Table \ref{tab:prompt_count} contains each set's total number of prompts used for KID and FID evaluations\footnote{\url{https://github.com/GaParmar/clean-fid}}.

\subsection{Iteration-based lambda scheduler}
There is a trade-off between unlearning and memory regularization while training to forget a class. This is because if we entirely focus on unlearning, it hinders the generalization over non-targeted concepts (as discussed in the Ablation study of Section \ref{sec:results}) and vice-versa. To mitigate this, we use iteration-based lambda scheduler for $\lambda_1$ and $\lambda_2$ with $\mathcal{L}_{\text{U}}$ and $\mathcal{L}_{pr}$ respectively in equation 8. This iteration-based scheduler is designed in a way that it gives more focus to $\mathcal{L}_{\text{U}}$ in the initial steps of training and more focus to $\mathcal{L}_{pr}$ in the final steps of training, resulting in a decreasing function to define $\lambda_1$ and an increasing function to define $\lambda_2$ based on total number of iterations used for training. Further, we first normalize the loss values of both $\mathcal{L}_{\text{U}}$ and $\mathcal{L}_{pr}$ and then use $\lambda_1$ and $\lambda_2$ resulting from this scheduler to compute the total loss value.

\begin{table*}[t]
\centering
\caption{Class-wise classification accuracies of Naive approach (left) and DUGE (right) for each set on our Dec-ImageNet-20 dataset. The diagonal values represent the accuracy for a target class to be unlearnt at step $\delta$. The original model is referred to as $\theta^0$. We observe that DUGE is successfully able to unlearn the target concepts decrementally while alleviating generalization erosion on targeted concepts.}
\resizebox{0.9\textwidth}{!}{%
\begin{tabular}{llccccc|ccccc}
\cline{2-12}
                       &            & \multicolumn{5}{c|}{Naive Approach}                                                                                                                                  & \multicolumn{5}{c}{DUGE (Proposed)}                                                                                                                                \\ \cline{3-12} 
                       &            & Apple                               & Broccoli                            & Backpack                           & Traffic Light                      & Umbrella       & Apple                              & Broccoli                           & Backpack                           & Traffic Light                      & Umbrella       \\ \cline{2-12} 
                       & $\theta^0$ & 85.44                               & 84.00                               & 97.42                              & 93.38                              & 98.85          & 85.44                              & 84.00                              & 97.42                              & 93.38                              & 98.85          \\ \cline{2-12} 
\multirow{5}{*}{\rotatebox[origin=c]{90}{Set 1}} & Step 1     & \multicolumn{1}{c|}{\textbf{11.00}} & 84.00                               & 89.22                              & 80.40                              & 90.00          & \multicolumn{1}{c|}{\textbf{8.00}} & 84.00                              & 96.00                              & 92.46                              & 98.81          \\ \cline{4-4} \cline{9-9}
                       & Step 2     & 0.00                                & \multicolumn{1}{c|}{\textbf{14.40}} & 76.35                              & 66.33                              & 84.16          & 6.43                               & \multicolumn{1}{c|}{\textbf{9.00}} & 96.00                              & 92.24                              & 98.85          \\ \cline{5-5} \cline{10-10}
                       & Step 3     & 9.40                                & 24.24                               & \multicolumn{1}{c|}{\textbf{8.20}} & 38.67                              & 58.93          & 5.82                               & 8.43                               & \multicolumn{1}{c|}{\textbf{2.00}} & 90.25                              & 94.87          \\ \cline{2-12} 
                       & Step 4     & 1.60                                & 5.60                                & 0.70                               & \multicolumn{1}{c|}{\textbf{9.92}} & 29.05          & 5.28                               & 5.00                               & 1.42                               & \multicolumn{1}{c|}{\textbf{8.27}} & 85.63          \\ \cline{7-7} \cline{12-12} 
                       & Step 5     & 4.50                                & 6.40                                & 12.62                              & 5.40                               & \textbf{10.80} & 5.18                               & 2.47                               & 0.45                               & 5.40                               & \textbf{10.82} \\ \cline{2-12} 
                       
\end{tabular}
}
\label{tab:set_1_results}
\vspace{8pt}
\centering
\resizebox{0.9\textwidth}{!}{%
\begin{tabular}{llccccc|ccccc}
\cline{2-12}
                       &            & \multicolumn{5}{c|}{Naive Approach}                                                                                                                                & \multicolumn{5}{c}{DUGE (Proposed)}                                                                                                                                 \\ \cline{3-12} 
                       &            & Goldfish                           & Bowtie                              & Harp                               & Parachute                          & Orange        & Goldfish                           & Bowtie                              & Harp                               & Parachute                          & Orange         \\ \cline{2-12} 
                       & $\theta^0$ & 99.42                              & 99.65                               & 99.27                              & 90.80                              & 99.00         & 99.42                              & 99.65                               & 99.27                              & 90.80                              & 99.00          \\ \cline{2-12} 
\multirow{5}{*}{\rotatebox[origin=c]{90}{Set 2}} & Step 1     & \multicolumn{1}{c|}{\textbf{0.29}} & 86.14                               & 65.59                              & 58.59                              & 64.20         & \multicolumn{1}{c|}{\textbf{1.62}} & 97.00                               & 99.00                              & 88.00                              & 98.80          \\ \cline{4-4} \cline{9-9}
                       & Step 2     & 1.28                               & \multicolumn{1}{c|}{\textbf{11.45}} & 65.28                              & 36.18                              & 50.02         & 1.62                               & \multicolumn{1}{c|}{\textbf{14.65}} & 98.82                              & 87.80                              & 96.15          \\ \cline{5-5} \cline{10-10}
                       & Step 3     & 1.44                               & 9.42                                & \multicolumn{1}{c|}{\textbf{8.00}} & 38.09                              & 52.87         & 1.62                               & 10.45                               & \multicolumn{1}{c|}{\textbf{5.25}} & 87.62                              & 90.66          \\ \cline{2-12} 
                       & Step 4     & 7.62                               & 24.42                               & 0.20                               & \multicolumn{1}{c|}{\textbf{1.68}} & 38.18         & 4.00                               & 8.81                                & 5.24                               & \multicolumn{1}{c|}{\textbf{5.63}} & 87.00          \\ \cline{7-7} \cline{12-12} 
                       & Step 5     & 1.24                               & 37.23                               & 1.65                               & 57.33                              & \textbf{9.65} & 4.00                               & 10.91                               & 5.20                               & 11.00                              & \textbf{12.00} \\ \cline{2-12} 
\end{tabular}
}
\label{tab:set_2_results}

\vspace{8pt}
\centering
\resizebox{0.9\textwidth}{!}{%
\begin{tabular}{llccccc|ccccc}
\cline{2-12}
                       &            & \multicolumn{5}{c|}{Naive Approach}                                                                                                                                                                                                                    & \multicolumn{5}{c}{DUGE (Proposed)}                                                                                                                                                                                                  \\ \cline{3-12} 
                       &            & \begin{tabular}[c]{@{}c@{}}Gas \\ Pump\end{tabular} & Sunglass                           & \begin{tabular}[c]{@{}c@{}}Computer\\ Mouse\end{tabular} & Broom                              & \begin{tabular}[c]{@{}c@{}}English \\ Springer\end{tabular} & Gas Pump                           & Sunglass                           & \begin{tabular}[c]{@{}c@{}}Computer\\ Mouse\end{tabular} & Broom                              & \begin{tabular}[c]{@{}c@{}}English\\ Springer\end{tabular} \\ \cline{2-12} 
                       & $\theta^0$ & 96.28                                               & 54.45                              & 98.85                                                    & 78.55                              & 99.00                                                       & 96.28                              & 54.45                              & 98.85                                                    & 78.55                              & 99.00                                                      \\ \cline{2-12} 
\multirow{5}{*}{\rotatebox[origin=c]{90}{Set 3}} & Step 1     & \multicolumn{1}{c|}{\textbf{8.0}}                   & 50.20                              & 81.00                                                    & 76.25                              & 82.12                                                       & \multicolumn{1}{c|}{\textbf{0.62}} & 54.45                              & 98.85                                                    & 78.55                              & 99.00                                                      \\ \cline{4-4} \cline{9-9}
                       & Step 2     & 1.48                                                & \multicolumn{1}{c|}{\textbf{8.82}} & 94.25                                                    & 58.45                              & 52.35                                                       & 0.48                               & \multicolumn{1}{c|}{\textbf{8.32}} & 95.48                                                    & 78.55                              & 99.0                                                       \\ \cline{5-5} \cline{10-10}
                       & Step 3     & 0.00                                                & 0.00                               & \multicolumn{1}{c|}{\textbf{0.00}}                       & 7.6                                & 25.82                                                       & 0.81                               & 4.08                               & \multicolumn{1}{c|}{\textbf{3.61}}                       & 78.50                              & 99.0                                                       \\ \cline{2-12} 
                       & Step 4     & 2.62                                                & 73.88                              & 32.81                                                    & \multicolumn{1}{c|}{\textbf{2.89}} & 20.53                                                       & 0.34                               & 4.10                               & 3.27                                                     & \multicolumn{1}{c|}{\textbf{5.63}} & 97.6                                                       \\ \cline{7-7} \cline{12-12} 
                       & Step 5     & 1.54                                                & 14.87                              & 15.42                                                    & 10.45                              & \textbf{0.23}                                               & 0.85                               & 0.24                               & 2.61                                                     & 3.62                               & \textbf{5.40}                                              \\ \cline{2-12} 
\end{tabular}
}
\label{tab:set_3_results}

\vspace{8pt}
\centering
\resizebox{0.9\textwidth}{!}{%
\begin{tabular}{llccccc|ccccc}
\cline{2-12}
                       &            & \multicolumn{5}{c|}{Naive Approach}                                                                                                                                                    & \multicolumn{5}{c}{DUGE (Proposed)}                                                                                                                                                      \\ \cline{3-12} 
                       &            & Golf Ball                          & Pizza                              & Tench                              & \begin{tabular}[c]{@{}c@{}}Garbage\\ Truck\end{tabular} & Teddy Bear    & Golf Ball                          & Pizza                              & Tench                              & \begin{tabular}[c]{@{}c@{}}Garbage \\ Truck\end{tabular} & Teddy Bear     \\ \cline{2-12} 
                       & $\theta^0$ & 95.80                              & 98.37                              & 71.40                              & 86.74                                                   & 100.00        & 95.80                              & 98.37                              & 71.40                              & 86.74                                                    & 100.00         \\ \cline{2-12} 
\multirow{5}{*}{\rotatebox[origin=c]{90}{Set 4}} & Step 1     & \multicolumn{1}{c|}{\textbf{0.88}} & 55.62                              & 16.34                              & 66.00                                                   & 88.78         & \multicolumn{1}{c|}{\textbf{9.79}} & 96.88                              & 71.40                              & 86.74                                                    & 100.00         \\ \cline{4-4} \cline{9-9}
                       & Step 2     & 0.00                               & \multicolumn{1}{c|}{\textbf{0.43}} & 9.40                               & 59.29                                                   & 87.90         & 7.83                               & \multicolumn{1}{c|}{\textbf{0.40}} & 66.88                              & 82.78                                                    & 100.00         \\ \cline{5-5} \cline{10-10}
                       & Step 3     & 3.33                               & 1.58                               & \multicolumn{1}{c|}{\textbf{0.00}} & 42.89                                                   & 85.63         & 8.40                               & 0.40                               & \multicolumn{1}{c|}{\textbf{8.88}} & 82.78                                                    & 99.60          \\ \cline{2-12} 
                       & Step 4     & 3.56                               & 1.54                               & 0.00                               & \multicolumn{1}{c|}{\textbf{1.30}}                      & 71.28         & 9.43                               & 0.04                               & 8.75                               & \multicolumn{1}{c|}{\textbf{1.67}}                       & 87.44          \\ \cline{7-7} \cline{12-12} 
                       & Step 5     & 1.60                               & 0.00                               & 0.00                               & 1.43                                                    & \textbf{4.03} & 9.28                               & 0.03                               & 8.64                               & 10.43                                                    & \textbf{15.18} \\ \cline{2-12} 
\end{tabular}
}
\label{tab:set_4_results}

\end{table*}

\begin{table*}[!t]
\scriptsize
\centering
\caption{The KID and FID values of all four sets computed between the original and $\delta$-step unlearnt models. Evaluation is performed on 5k prompts from the validation set of the COCO dataset. We observe that DUGE successfully maintains the generalization capability of the original models up to the last decremental step, while the baseline method suffers from generalization erosion starting from the second step.}
\resizebox{\textwidth}{!}{%
\begin{tabular}{lcccccccccccccccc}
\hline
       & \multicolumn{4}{c}{Set 1}                                   & \multicolumn{4}{c}{Set 2}                                   & \multicolumn{4}{c}{Set 3} &
       \multicolumn{4}{c}{Set 4} \\ \cline{2-17} 
       & \multicolumn{2}{c}{Baseline} & \multicolumn{2}{c}{Proposed} & \multicolumn{2}{c}{Baseline} & \multicolumn{2}{c}{Proposed} & \multicolumn{2}{c}{Baseline} & \multicolumn{2}{c}{Proposed}& \multicolumn{2}{c}{Baseline}  & \multicolumn{2}{c}{Proposed}\\ \cline{2-17} 
       & KID            & FID         & KID            & FID         & KID            & FID         & KID            & FID         & KID            & FID         & KID            & FID & KID            & FID & KID            & FID         \\ \hline
Step 1 & 1.79e-3        & 7.01        & 1.98e-5        & 2.44        & 3.74e-3        & 12.95       & 7.87e-6        & 2.10        & 5.12e-3        & 18.05       & 8.40e-5        & 2.75      & 1.92e-3        & 9.34        & 2.08e-6        & 2.25   \\
Step 2 & 3.65e-3        & 13.01       & 3.49e-5        & 2.55        & 3.08e-3        & 10.37       & 3.95e-5        & 2.52        & 3.56e-3        & 12.78       & 1.07e-4        & 2.91    & 3.73e-3        & 15.22       & 3.14e-5        & 2.84    \\
Step 3 & 3.17e-3        & 11.19       & 1.07e-4        & 3.31        & 4.37e-3        & 14.19       & 3.99e-5        & 2.67        & 0.01           & 47.51       & 1.10e-4        & 2.79  & 7.74e-3        & 18.97       & 3.35e-5        & 2.97      \\ \hline
Step 4 & 7.61e-3        & 24.24       & 1.65e-4        & 3.41        & 4.30e-3        & 16.77       & 9.47e-5        & 2.90        & 5.31e-3        & 21.43       & 1.29e-4        & 3.14 & 1.14e-2        & 24.19       & 1.83e-4        & 3.46       \\
Step 5 & 4.54e-3        & 14.12       & 1.97e-4        & 3.82        & 5.28e-3        & 18.63       & 3.01e-4        & 3.64        & 0.01           & 32.97       & 1.66e-4        & 3.20   & 1.44e-2        & 36.42       & 5.66e-4        & 3.72     \\ \hline
\end{tabular}
}
\label{tab:kid_fid}
\end{table*}

\section{Results and Discussion}
\label{sec:results}
In this section, we analyze DUGE's performance from unlearning for a target concept and generalization over non-targeted concepts. Since previous work on decremental learning does not exist, we implemented a Naive Finetuning method as a baseline approach for comparison. As described earlier, in this method, the model is iteratively fine-tuned to map random images with prompts describing the target concepts. The details of these mappings are provided in Table \ref{tab:replacement}. The replacement class for each target class is randomly chosen from the ImageNet dataset. The Naive Approach and DUGE are evaluated using class-wise accuracy, FID, and KID at each decremental step. In addition to these four sets, we also present evaluation results of DUGE with decremental unlearning setting~on: 
\begin{itemize}
    \item Closely related concepts to analyze how DUGE performs on correlated concepts.
    \item Removing identities, for this, we utilize the concept Bench Identity Set introduced by Zhang et al. \citep{zhang2024forget}.
    \item Removing artistic styles, for this we present qualitative results of DUGE on removing three artistic styles.
\end{itemize}

Next, we present both quantitative and qualitative analyses of the proposed DUGE algorithm: 

\noindent\textbf{Quantitative Analysis:} For unlearning, the class-wise classification accuracy is reported in Table \ref{tab:set_1_results} for all sets of our Dec-ImageNet-20 dataset. For the decremental setup, we present the results in a 5x5 matrix (excluding original classification accuracy) where diagonal values represent the unlearning classification accuracy for the target class at step $\delta$; therefore, the lower, the better. For the Naive approach, we observe that though lower diagonal values for all sets represent good unlearning performance; however, it is not stable for previous unlearnt classes. This is made evident by a rise in class-wise accuracy for decremental steps 3 and 5 for Set 1 (Table \ref{tab:set_1_results}); steps 2, 4, and 5 for Set 2 (Table \ref{tab:set_2_results}) and steps 4 and 5 for Set 3 (Table \ref{tab:set_3_results}). 

Moreover, we observe a significant drop in accuracy for classes yet to be unlearnt starting from Step 2 only for the naive approach for Sets 1 and 3; and from Step 1 for Set 2. This also highlights a poor generalization of the Naive approach for decremental setup. From the diagonal values in the tables, we notice a significant drop in classifier performance, implying that DUGE is able to forget the target classes for all sets. For all decremental steps, DUGE maintains the unlearning performance of previous unlearnt classes, except for the $5^{th}$ decremental step for the orange class (Table \ref{tab:set_2_results}). We also notice that DUGE is robust to generalization erosion and maintains the original classification accuracy for classes that are not unlearnt up to three decremental steps. However, the performance reduces for the $5^{th}$ decremental step. 

To further test generalization erosion, the KID and FID values for the three sets are presented in Table \ref{tab:kid_fid}. Compared to the naive approach, the lower values of DUGE clearly show that it can generalize better on the prompts from the MS-COCO dataset. However, we see a linear increase in both values for all sets when learned decrementally for subsequent classes. This happens because, after each unlearning step, the model weights deviate away from the original model, resulting in a slight degradation in the quality of the generated output.

\begin{figure}[t]
\centering
    \includegraphics[scale=0.31]{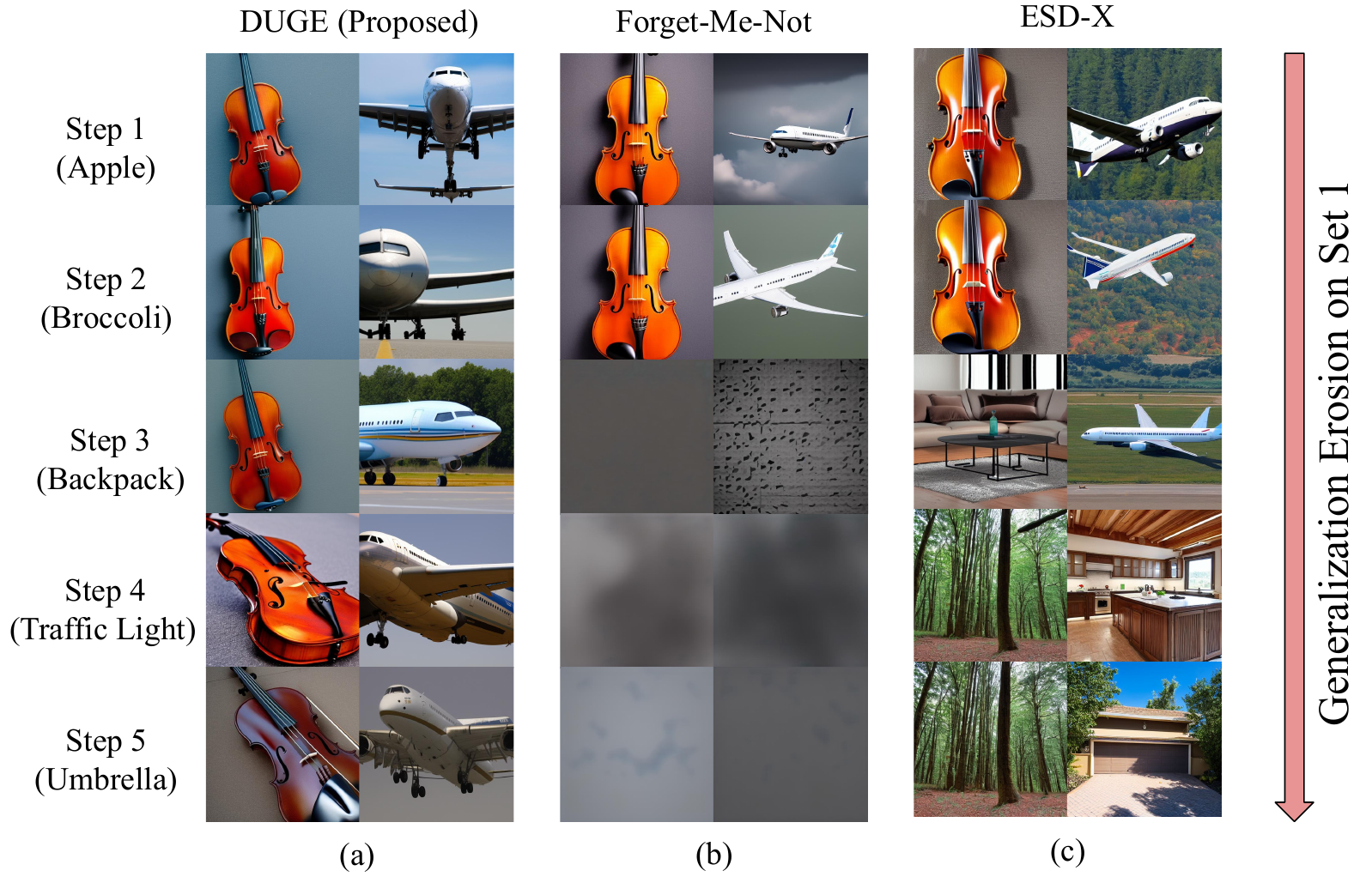}
    \caption{Visualizing the generation results on set 1 for the proposed algorithm DUGE, alongside FMN \citep{zhang2024forget} and ESD-X \citep{gandikota2023erasing}. It is evident that \citep{zhang2024forget} and \citep{gandikota2023erasing} suffer from generalization erosion in a continual unlearning setting when prompted with ‘an image of a violin’ and ‘an image of an airplane,’ whereas DUGE effectively generates samples at each decremental step.}
\label{fig:comparison_algos}
\end{figure}

\begin{figure*}[t]
\centering
    \includegraphics[scale=0.38]{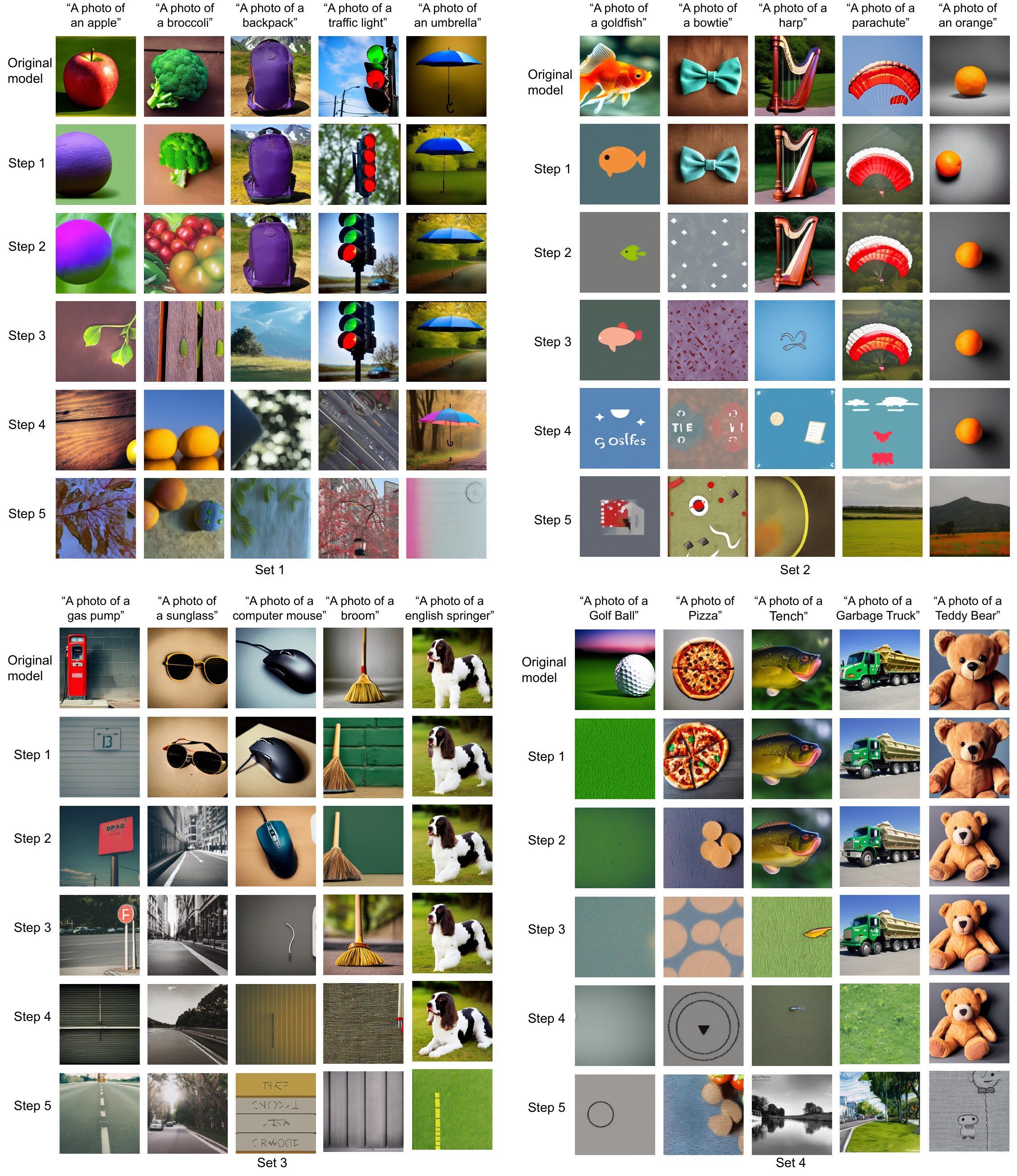}
    \caption{Qualitative results from various sets, illustrating samples of unlearnt and other classes at each decremental step $\delta$. It is evident that across all sets, DUGE consistently unlearns the target class while preserving the knowledge of other concepts.}
\label{fig:qualitative_results_sets}
\end{figure*}

\begin{figure*}[t]
\centering
    \includegraphics[scale=0.42]{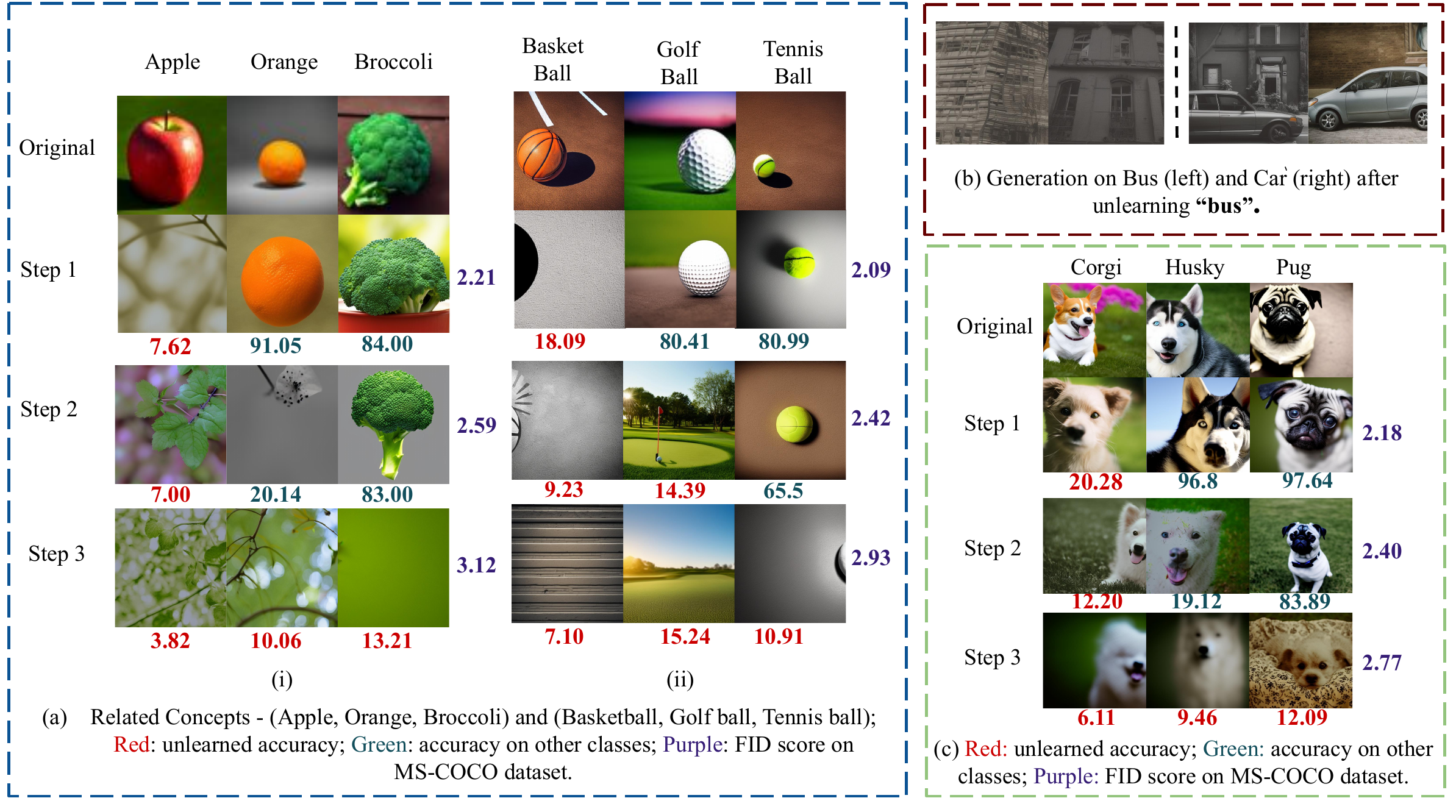}
    \caption{Illustrating the performance of DUGE on correlated concepts. (a) Continual unlearning of concepts like Apple, Orange, and Broccoli, with performance visualized across all concepts. (b) Visualization of DUGE’s performance on correlated concepts such as car and bus. (c) Performance of DUGE on Concept Bench \citep{zhang2024forget} for different dog breeds. We observe that DUGE successfully decrementally unlearns correlated concepts, including (a) fruits and vegetables, and various types of balls; (b) bus and car; (c) different dog breeds.}
\label{fig:correlated_concepts}
\end{figure*}

\begin{figure}[t]
\centering
    \includegraphics[scale=0.34]{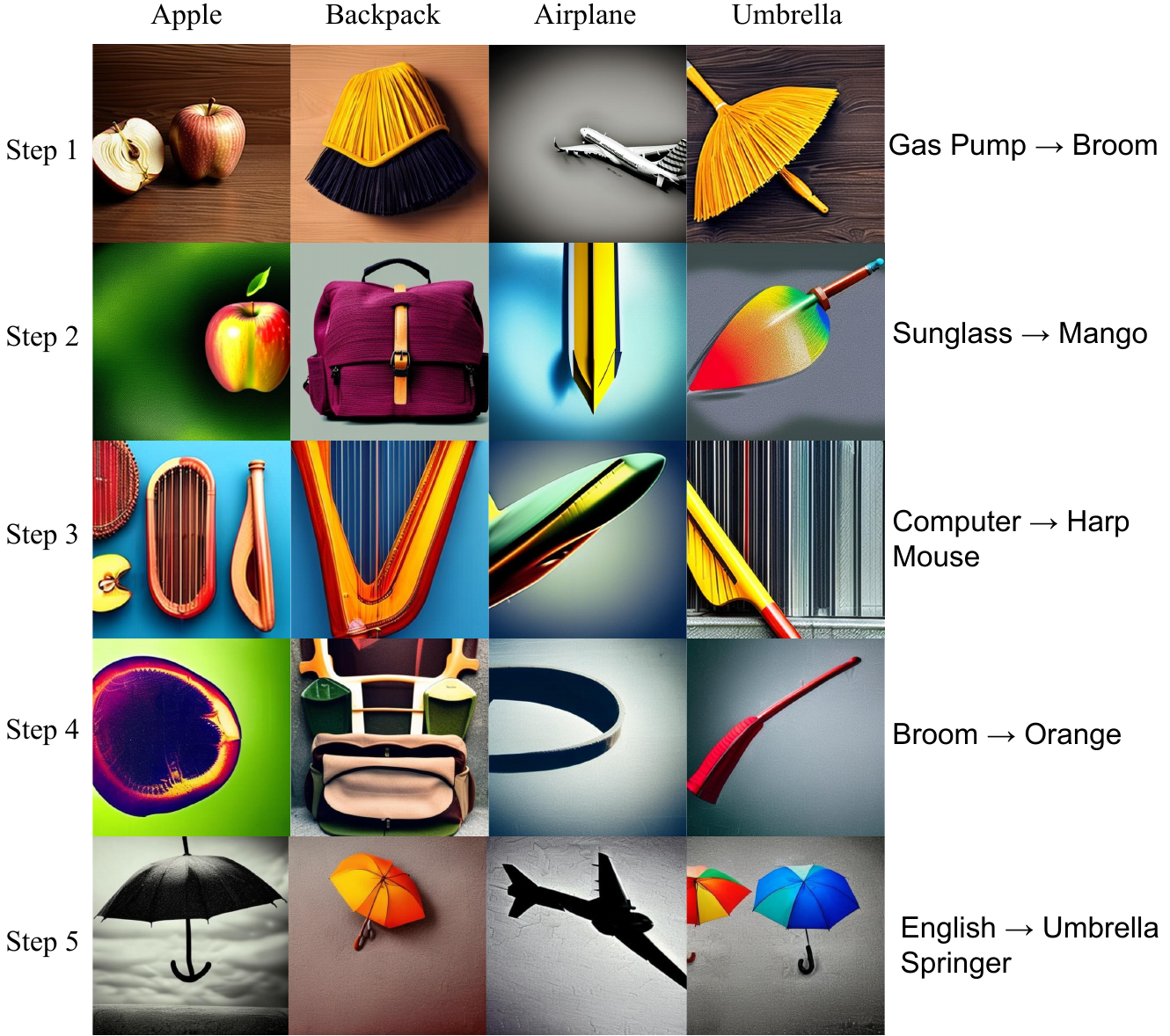}
    \caption{Samples generated from the Naive-Finetuned model, unlearnt for non-targeted classes of Set 3. This shows that Naive fine-tuning for decremental unlearning suffers from generalization erosion and binds non-targeted classes with replacement classes.}
\label{fig:generalization_erosion}
\end{figure}

\noindent\textbf{Qualitative Analysis:} 
Figure \ref{fig:qualitative_results_sets} showcases the output for all classes of each set. The results in the first row are generated by the original/source model, followed by the samples from each decremental step. It can be observed that DUGE successfully forgets the concepts for each decremental step while ensuring the preservation of high visual quality for the other concepts. For the naive approach, guided by the quantitative results, we report a few samples of non-targeted concepts generated by the unlearnt baseline model on Set 3 in Figure \ref{fig:generalization_erosion}. We observe poor generalizability of the baseline approach as it tends to affect non-target concepts. For instance, it can be seen that when the gas pump class is bound with the broom class, the model tends to bind broom images for unrelated classes. This can be seen for all decremental steps from the baseline approach, giving rise to generalization erosion in decremental learning.

\noindent\textbf{Correlated Concepts:} It is an obvious assumption that unlearning a concept might affect the performance of the model on closely correlated concepts. We test DUGE on such correlated concepts and evaluate it both qualitatively and quantitatively. In the first experiment, we consider unlearning fruits and vegetables, i.e., apples, followed by oranges, and then broccoli. From Fig. \ref{fig:correlated_concepts} (a) (i), we observe that DUGE is successfully able to unlearn these concepts decrementally. We test DUGE to decrementally unlearn different kinds of balls, i.e., unlearning basketball, followed by golf ball, and finally tennis ball. From Fig. \ref{fig:correlated_concepts} (a) (ii), we observe that the model is successfully able to unlearn basketball while retaining the performance on golf ball and tennis ball. 

\par In step 2, the model forgets the concept of the golf ball and retain the concept of a tennis ball. The challenge of incrementally unlearning correlated concepts is emphasized by the low classification performance of 14.39\% for golf balls and a moderate 65.5\% for tennis balls. Despite this, DUGE demonstrates its capability to incrementally unlearn these concepts. In Fig. \ref{fig:correlated_concepts} (b), we also observe successful unlearning of the concept ‘bus’ while maintaining its performance on the concept ‘car’, as they are both vehicles and are correlated concepts. This demonstrates the efficacy of DUGE on correlated concepts.

\noindent\textbf{Comparison with Existing Algorithms}
Recent studies from Gandikota et al. \citep{gandikota2023erasing} and Zhang et al. \citep{zhang2024forget} introduced the concept of unlearning in text-to-image foundational models. We evaluate their performance on generalization erosion using concepts from Set 1 of the Dec-ImageNet-10 dataset and compare them with our proposed method. Each algorithm was tested with the prompts "An image of a violin" and "A photo of an airplane" after unlearning each concept from Set 1. 

As shown in Fig. \ref{fig:comparison_algos} (b), Forget-Me-Not begins to experience generalization erosion immediately after the second step. Similarly, ESD-X\footnote{
ESD-X utilizes SD 1.4 in its implementation} (Fig. \ref{fig:comparison_algos} (c)) loses its ability to generalize the concept of a violin after the second step and an airplane after the third step. In contrast, our proposed algorithm, DUGE, appears to be entirely resistant to generalization erosion, even when decrementally unlearning all concepts from Set 1. This finding suggests that current algorithms are not yet equipped for decremental unlearning, and DUGE serves to fill this gap.

\noindent\textbf{Concept Bench:} We present the performance of DUGE on animal subcategories described in Concept Bench for decremental unlearning, namely: Corgi, Husky, and Pug which belong to the general category of dog. The performance is visualized in Fig. \ref{fig:correlated_concepts} (c). We clearly observe that after forgetting Corgi, the model is able to generalize over Husky and Pug. Moreover, after forgetting Husky, the model still retains pug. At the third decremental step, the model has successfully unlearned all the breeds. We also notice that DUGE only forgets the feature corresponding to each breed while still generating dogs as output. This shows the robustness of DUGE on Concept Bench for continual unlearning.

\begin{table*}[t]
\centering
\caption{Class-wise classification accuracies for two ablation experiments performed - one with just unlearning loss $\mathcal{L}_U$ and other with unlearning loss along with memory regularization loss $\mathcal{L}_U + \mathcal{R}_{pr}$. Diagonal values represent the classification accuracy for a target class to be unlearnt at step $\delta$. The original model is referred as $\theta^0$. Lower accuracy values for non-target classes in each step shows the significance of incorporating memory regularization along with unlearning and weight-regularization loss. }
\resizebox{0.9\textwidth}{!}{%
\begin{tabular}{lccccc|ccccc}
\hline
  & \multicolumn{5}{c|}{$\mathcal{L}_U$}         & \multicolumn{5}{c}{$\mathcal{L}_U + \mathcal{R}_{pr}$}                      \\ \cline{2-11} 
& Apple                               & Broccoli                            & Backpack                           & \begin{tabular}[c]{@{}c@{}}Traffic Light\end{tabular} & Umbrella       & Apple                              & Broccoli                           & Backpack                           & \begin{tabular}[c]{@{}c@{}}Traffic Light\end{tabular} & Umbrella       \\ \hline
$\theta^0$ & 85.44                               & 84.00                               & 97.42                              & 93.38                                                   & 98.85          & 85.44                              & 84.00                              & 97.42                              & 93.38                                                   & 98.85          \\ \hline
Step 1                                                    & \multicolumn{1}{c|}{\textbf{14.00}} & 84.02                               & 86.78                              & 80.60                                                   & 97.4          & \multicolumn{1}{c|}{\textbf{13.44}} & 84.00                              & 89.97                              & 83.26                                                   & 97.80          \\ \cline{3-3} \cline{8-8}
Step 2                                                    & 6.21                               & \multicolumn{1}{c|}{\textbf{0.80}} & 48.09                              & 47.21                                                   & 80.79          & 5.01                               & \multicolumn{1}{c|}{\textbf{1.80}} & 48.09                              & 49.38                                                   & 82.77         \\ \cline{4-4} \cline{9-9}
Step 3                                                    & 0.80                                & 0.40                              & \multicolumn{1}{c|}{\textbf{0.0}} & 1.80                                                   & 8.76          & 0.40                               & 0.20                               & \multicolumn{1}{c|}{\textbf{0.00}} & 1.80                                                  & 11.02          \\ \hline
Step 4                                                    & 0.00                               & 0.00                                & 0.00                               & \multicolumn{1}{c|}{\textbf{0.00}}                      & 0.00          & 0.00                             & 0.00                               & 0.00                              & \multicolumn{1}{c|}{\textbf{0.00}}                      & 0.00          \\ \cline{6-6} \cline{11-11} 
Step 5                                                    & 0.00                                & 0.00                                & 0.00                              & 0.00                                                   & \textbf{0.00} & 0.00                              & 0.00                               & 0.00                               & 0.00                                                    & \textbf{0.00} \\ \hline
\end{tabular}
}
\label{tab:ablation_results}
\end{table*}

\begin{figure}[t]
\centering
    \includegraphics[scale=0.52]{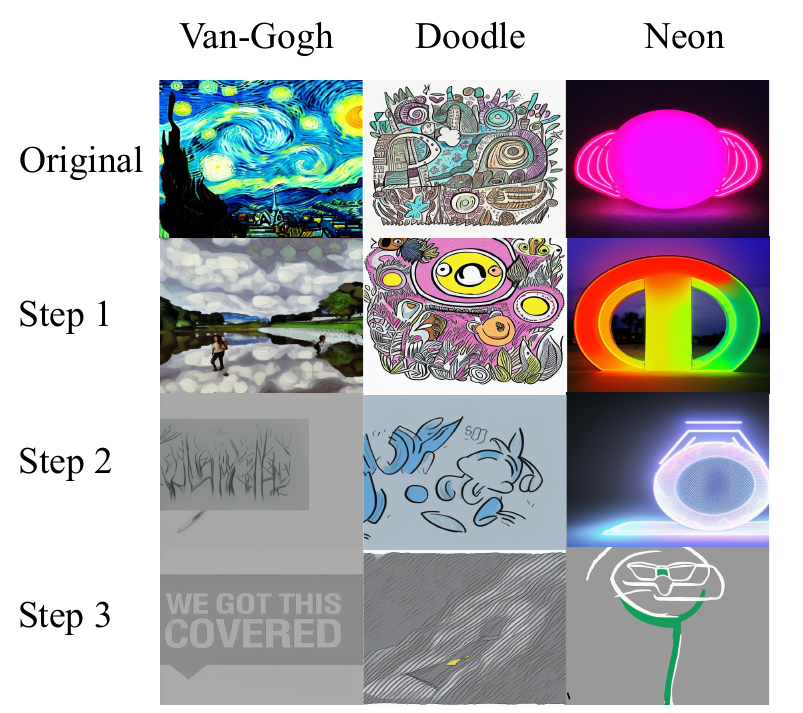}
    \caption{Visualization of output generated by DUGE after decrementally unlearning different artistic styles demonstrating the performance on practical applications.}
\label{fig:artistic_styles}
\end{figure}

\noindent\textbf{Performance of DUGE on Practical Applications:}
This paper focuses on establishing the theoretical foundation and demonstrating the feasibility of DUGE in decremental unlearning. However, a more comprehensive exploration of practical applications is necessary to suggest the practical deployability of the proposed method. In this experiment, we explore the practical applications of DUGE and test it in real-world scenarios, such as the removal of artistic styles at the individual request of artists. The efficacy of DUGE in practical applications can be visualized in Fig. \ref{fig:artistic_styles}, which shows the decremental removal of artistic styles. We tested DUGE on Van Gogh, followed by Doodle, and finally Neon, observing that DUGE is able to successfully unlearn different artistic styles incrementally. This demonstrates the practical deployment of DUGE in real-world applications.

\begin{table}[]
\centering
\caption{The KID and FID values computed for two ablation settings: the first (left) is with unlearning loss, and the second (right) is unlearning with weight-regularization loss. The higher values of both settings from step 2 shows the significance of incorporating memory along with weight regularization and unlearning loss.}
\begin{tabular}{lcccc}
\hline
       & \multicolumn{2}{c}{$\mathcal{L}_U$} & \multicolumn{2}{c}{$\mathcal{L}_U + \mathcal{R}_{pr}$} \\ \cline{2-5} 
       & KID          & FID        & KID          & FID        \\ \hline
Step 1 & 0.00028                 & 3.83576                 & 0.00027                 & 3.76963                 \\
Step 2 & 0.00165                 & 8.36439                 & 0.00150                 & 8.14377                 \\
Step 3 & 0.02630                 & 76.2159                 & 0.02388                 & 66.15763                \\ \hline
Step 4 & 0.20722                 & 250.74670               & 0.21080                 & 248.52227               \\
Step 5 & 0.31114                 & 347.02307               & 0.32166                 & 351.43423               \\ \hline
\end{tabular}
\label{tab:ablation_quantitative}
\end{table}

\begin{figure*}[!t]
\centering
    \includegraphics[scale=0.475]{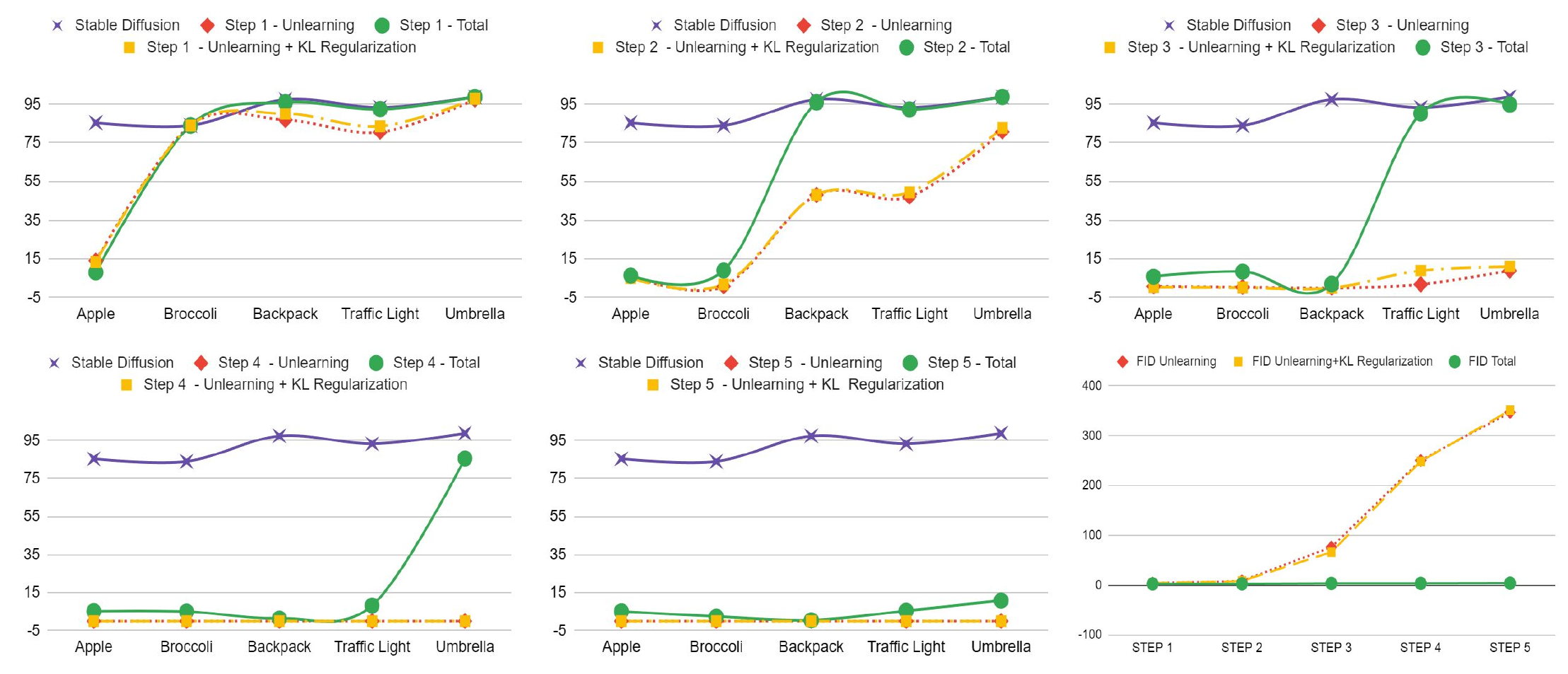}
    \caption{Graphs displaying classification accuracies for ablation experiments conducted over 5 decremental steps on Set 1 (left to right, then top to bottom), followed by a graph showing FID values for each decremental step. The first five graphs compare three settings: unlearning loss, total proposed loss, and unlearning with KL regularization. It is evident that using only the unlearning loss results in a drop in accuracies for other classes from step 2 onwards, while the proposed total loss performs better in each decremental setup. A similar trend is observed in the last graph for generalization capabilities, as indicated by FID values for each decremental setup.}
    
    
\label{fig:ablation_graphs}
\end{figure*}

\begin{figure}[t]
        \includegraphics[scale=0.31]{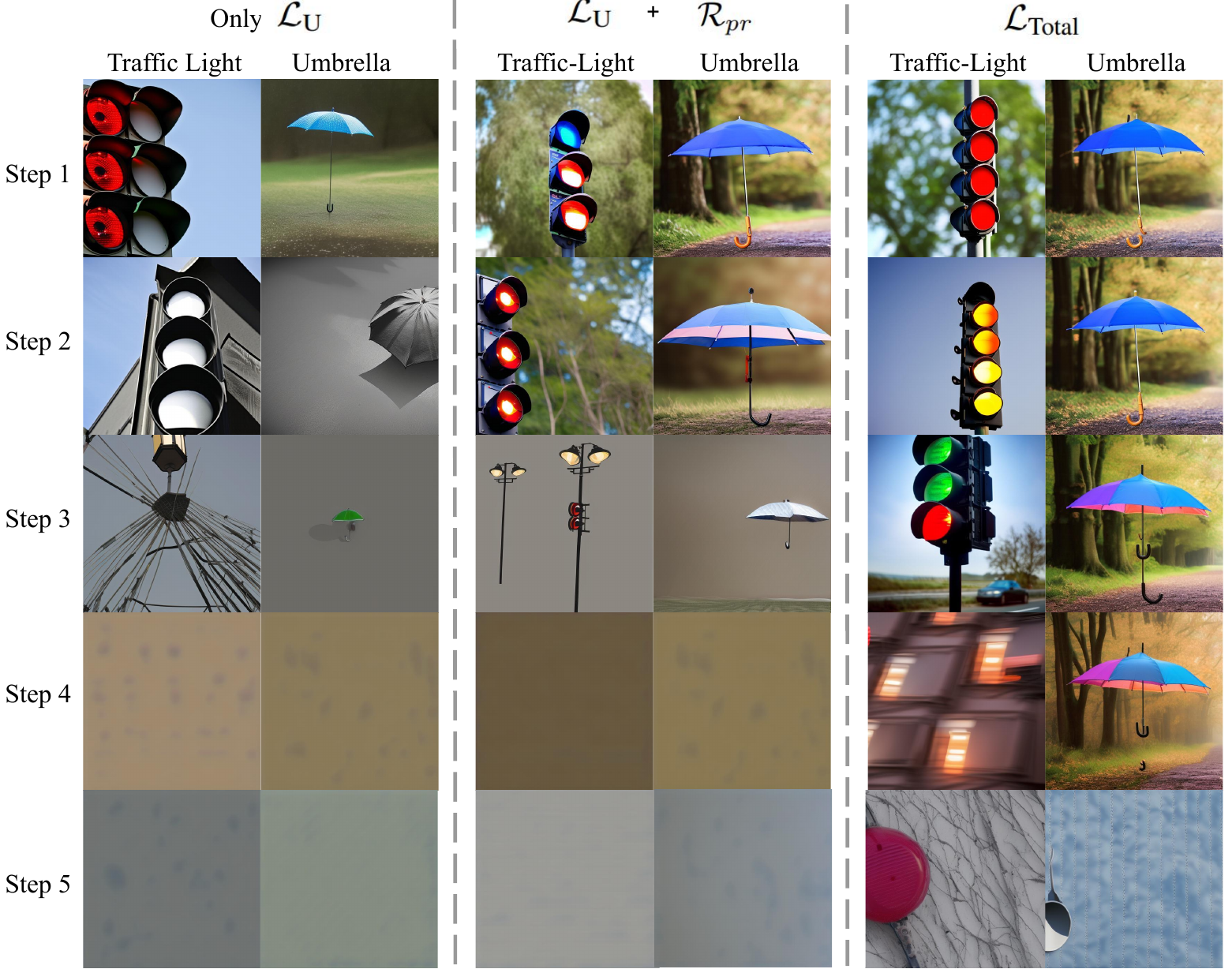}
    \caption{Visualization of three ablation settings for the classes Traffic Light and Umbrella in each decremental setup. It is evident that the performance of the first two settings is affected from step 2 onwards, while the proposed model maintains the generation of these two classes until they are unlearned.}
\label{fig:ablation}
\end{figure}

\noindent\textbf{User Study:} 
To gain a human perspective on the performance of the proposed DUGE algorithm, we conducted a \textit{User Study} involving 15 participants (ages ranging from 18 to 65 years, consisting of seven males and eight females). These participants were presented with 25 images generated from set 1 (5 images from each class) of Dec-ImageNet-20. Each participant was asked to assign a score of 0 if the target object was present in the image and 1 if not. In cases of uncertainty, a rating of 0.5 could be given. These scores were then accumulated, and an average score was computed for all participants. Our human evaluation yielded an average score of 95.6\%, indicating that the generated output successfully unlearned all the target concepts.

\noindent\textbf{Ablation Study:}
We also conducted an ablation study for the three loss functions in our proposed formulation. We first begin with only unlearning loss $\mathcal{L}_U$, followed by the addition of weight-regularization $\mathcal{R}_{pr}$. Finally, we also utilize prior-preservation loss $\mathcal{L}_{pr}$ also to understand the effect of each loss component in decremental unlearning. The ablation experiments are performed on Set 1 of the Dec-ImageNet-20 dataset and classification accuracy with FID is computed for all classes over each decremental step. The results are reported in Figure \ref{fig:ablation_graphs}. From the first 5 graphs, we observe that when only $\mathbf{L}_U$ is employed for decremental learning, it fails to generalize after $2^{nd}$ decremental step and degrades the performance over non-targeted concepts. This is evidenced by classification accuracies over unlearnt classes and the FID score available in the last graph. Adding regularization loss to the unlearning loss also results in similar behavior for more than $2$ decremental steps. Still, a slight improvement can be seen in the classification accuracies for unlearnt classes in decremental set 1. 

\par Figure \ref{fig:ablation} also supports this finding, and we can see very poor generalizability of $\mathcal{L}_U$ and $\mathcal{L}_U + \mathcal{R}_{pr}$ scenarios starting from decremental step 3. We can clearly conclude that although the $\mathcal{L}_U$ can unlearn a single target class well while maintaining generalization, it cannot handle unlearning in a decremental fashion. Therefore, adding a small memory and corresponding prior preservation loss $\mathcal{L}_{pr}$ helps to prevent generalization erosion in decremental unlearning.

Table \ref{tab:ablation_results} and Table \ref{tab:ablation_quantitative} present the numerical quantification of the ablation study conducted. From Table \ref{tab:ablation_results}, it can be observed that while DUGE, without memory regularization, is effective in unlearning a target class, this comes at the expense of weaker generalization performance. It can also be observed that DUGE without memory regularization can be utilized for just a single decremental step without hindering the generalizability of the source model. However, for additional decremantal steps neither $\mathcal{L}_U$ nor $\mathcal{L}_U + \mathcal{R}_{pr}$ is capable of performing well, as evidenced by the generalization erosion highlighted in both Table \ref{tab:ablation_results} and~Table \ref{tab:ablation_quantitative}.

\subsection{Computational Runtime}

We calculate the time required to unlearn a single concept using our proposed algorithm. During this process, we unlearn five distinct concepts with $m = 50$. Notably, the average time to unlearn each concept was clocked at 25 minutes. These measurements were obtained using a single Nvidia A100 GPU.

\subsection{Limitations}
\noindent We note the following limitations of the proposed \textit{DUGE} algorithm when used for unlearning or decremental learning in foundational generative models:
\begin{itemize}
   \item The algorithm proposed is an intervention in the cross-attention mechanism, and as such, it can only be implemented in generative models that utilize this mechanism. Therefore, its application is confined to text-to-image diffusion models. In its current form, it cannot be applied to Generative Adversarial Networks (GANs) and Variational Autoencoders (VAEs).
   \item Although we can efficiently unlearn numerous concepts across a broad spectrum, this does not guarantee that the target concept cannot be produced through a different, unrelated text prompt. For a model that has undergone several decremental steps, a notorious user might still be able to generate traces of an unlearned concept through inventive/engineered prompts, despite the concept having been decrementally forgotten from the model.
   
\end{itemize}

\section{Conclusion}

This research presents a new paradigm termed continual unlearning in foundational text-to-image generative models, focusing on the selective removal of specific concepts like copyrighted content or artistic styles. To the best of our knowledge, this is the first work to propose continual unlearning in text-to-image generative models while also addressing the issue of generalization erosion. To address this challenge, we introduce DUGE, a method that uses cross-attention maps and memory regularization to maintain the model’s overall capabilities while removing targeted knowledge. When tested on the Dec-ImageNet-20 dataset, DUGE effectively eradicates multiple concepts without compromising the model’s capacity to generate unrelated images. It also demonstrates resistance to generalization erosion, a common issue where quality and diversity typically decline after unlearning. 

\section{Acknowledgement}
The authors thank all volunteers in the user study. This research is supported by the IndiaAI mission and Thakral received partial funding through the PMRF~Fellowship.
{
    \small
    \bibliographystyle{ieeenat_fullname}
    \bibliography{main}
}


\end{document}